**Cluster-based human-in-the-loop strategy for improving machine learning-based circulating tumor cell detection in liquid biopsy**

Hümeyra Husseini-Wüsthoff[1,2,3,*], Sabine Riethdorf[4], Andreas Schneeweiss[5], Andreas Trumpp[6], Klaus Pantel[4], Harriet Wikman[4], Maximilian Nielsen[1,2,3,7], and René Werner[1,2,3,7]

[1] Institute for Applied Medical Informatics, University Medical Center Hamburg-Eppendorf, Hamburg, Germany
[2] Department of Computational Neuroscience, University Medical Center Hamburg-Eppendorf, Hamburg, Germany
[3] Center for Biomedical Artificial Intelligence (bAIome), University Medical Center Hamburg-Eppendorf, Hamburg, Germany
[4] Institute of Tumor Biology, University Medical Center Hamburg-Eppendorf, Hamburg, Germany
[5] National Center for Tumor Diseases, Heidelberg University Hospital and German Cancer Research Center, Heidelberg, Germany
[6] Division of Stem Cells and Cancer, German Cancer Research Center (DKFZ) and DKFZ-ZMBH Alliance, Heidelberg, Germany
[7] Senior authors
[*] Correspondence: h.husseini-wuesthoff@uke.de

# Summary

Detection and differentiation of circulating tumor cells (CTCs) and non-CTCs in blood draws of cancer patients pose multiple challenges. While the gold standard relies on tedious manual evaluation of an automatically generated selection of images, machine learning (ML) techniques offer the potential to automate these processes. However, human assessment remains indispensable when the ML system arrives at uncertain or wrong decisions due to an insufficient set of labeled training data. This study introduces a human-in-the-loop (HiL) strategy for improving ML-based CTC detection. We combine self-supervised deep learning and a conventional ML-based classifier and propose iterative targeted sampling and labeling of new unlabeled training samples by human experts. The sampling strategy is based on the classification performance of local latent space clusters. The advantages of the proposed approach compared to naive random sampling are demonstrated for liquid biopsy data from patients with metastatic breast cancer.

# Keywords

# Introduction

Continuous research on cancer over the last decades has led to a steady improvement in early detection and treatment, resulting in an increase in patient outcomes in terms of both survival rates and quality-adjusted life years[1–5]. Thus, monitoring cancer progression is important to evaluate individual treatment responses. Especially detection of metastasis is of high interest as it is the driving force behind progression and strongly correlates with patient outcome. As part of the metastatic cascade, tumor cells disseminate from the primary tumor and circulate primarily through the bloodstream to surrounding or distant organs[6]. These cells are referred to as circulating tumor cells (CTCs). Numerous studies have analyzed blood draws of patients to investigate the spread of tumor cells and to identify related markers in liquid biopsy (LB)[7–9]. However, challenges arise from the heterogeneity of CTCs, such as various phenotypic expressions[6], CTC rarity (< 10 cells $ml^{-1}$)[10], and reliable CTC detection is still associated with difficulties.

So far, only one solution is cleared for routine clinical analysis of CTCs from metastatic breast, prostate and colorectal cancers[11] by the U.S. Food and Drug Administration (FDA)[12]: the CellSearch® (CS) system (Menarini Silicon Biosystems, Bologna, Italy). Multiple clinical studies with the CS system have demonstrated a tight correlation between CTC appearance and poor prognosis in metastatic breast cancer[13–15]. CS follows a three-step process. First, in the Autoprep system (Menarini), blood samples are processed with a widely used method for isolating CTCs from the bulk of blood cells through EpCAM (Epithelial cell adhesion molecule)-based immunomagnetic separation. Secondly, these enriched cells are fluorescently labeled. Subsequently, the cells are placed in a magnetic cartridge where a magnetic force draws them to a single focal depth[16]. The cartridge is then transferred into the Autoanalyzer (Menarini) for automated microscopy scanning. In the next step, images containing positive signals in the 4',6-diamidino-2-phenylindole, dihydrochloride (DAPI) and phycoerythrin (PE) channels in close proximity[16] are automatically selected and presented in an image gallery by a software. Finally, all presented images have to be evaluated manually by a trained operator to identify CTCs and differentiate them from contaminating leukocytes or artifacts[12] according to defined criteria. A cell is considered a CTC when it has a round or oval shape with a diameter of at least 4 µm, a DAPI-positive nucleus, cytoplasmic PE staining as indicator for keratin positivity, but no APC staining to exclude CD45-positive leukocytes.

Although human assessment of CTC candidate images will remain necessary, there is a strong need for greater automation of CTC detection and analysis. Zeune et al.[17] used LB data from various cancer entities, a.o. metastatic breast cancer, in a supervised deep learning (DL) approach. The cell images were sampled from cartridge images acquired by CS using the ACCEPT tool[18], and automatically generated annotations were manually corrected by human experts. They further shed light on the model behavior by investigating the latent space using dimension reduction and analyzing clustering behavior for different sub-populations of cells. Building on the findings of Zeune et al.[17], Nanou et al.[19] presented a strategy for semi-supervised labeling of training data by utilizing a latent space analysis and identifying additional unambiguous samples in dense CTC and non-CTC regions identified by a k-nearest neighbor (kNN)-based analysis[19]. In parallel, self-supervised learning has advanced in the medical field and showed promising performance utilizing less annotated data but taking advantage of the availability of often large amounts of unlabeled samples[20]. For example, Husseini et al.[21] demonstrated that a self-supervised setup for CTC detection in a breast cancer cohort outperforms supervised approaches with only a fraction of the annotations needed.

In this work and with the focus on differentiation of CTC and non-CTC images, we aim to bridge the gap between self-supervised and semi-supervised approaches with the introduction of a novel human-in-the-loop (HiL) strategy for providing additional meaningful training samples to an initial classifier. While Nanou et al.[19] sampled from dense CTC and non-CTC latent space areas to increase the certainty of automatically pseudo-labeled data points, we hypothesize that sampling from areas with higher uncertainty is more beneficial since these areas are where most of the false classifications take place. This is intrinsically not accounted for by Nanou et al.[19]; for their approach, the classification of cells within these regions remains uncertain. To enable rapid classifier adjustment upon the availability of new training samples, we combine a custom self-supervised (self-*di*stillation with *no* labels, DINO[22]) pretrained image encoder with a lightweight machine learning classifier (support vector machine, SVM) following the setup by Husseini et al.[21] and Nielsen et al.[20]. The study builds on CS cartridge images from 90 metastatic breast cancer patients. The proposed framework deploys the StarDist algorithm to extract single-cell crops from the cartridge images[23]. For cell classification, we combine both the advantages of state-of-the-art machine learning and human experience and intervention. We demonstrate the feasibility of the HiL strategy by experiments based on both simulations (simulated HiL) and with a human operator in the loop (real-world HiL). Our major contributions and major findings are:

1) *Detailed latent space analysis*: We provide a detailed analysis of the latent space cell representations for metastatic breast cancer LB data and demonstrate that clusters with differing classification performance exist in the latent space.
2) *Proposal of an efficient HiL strategy:* Based on finding (1), we introduce a novel iterative, local classifier performance-driven sampling and labeling strategy and demonstrate its feasibility and effectiveness.
3) *Public availability of training framework and models*: The complete framework (2), including model weights for the image encoder and a pipeline to generate cell images from cartridge ones, is made publicly available.

# Results

## Framework overview and the HiL principle

The proposed framework consists of three main modules (Figure 1): single cell image extraction (Figure 1A); self-supervised image encoder training using unlabeled cell images (Figure 1B); and cell classification based on the HiL principle (Figure 1A, lower part), utilizing a cluster analysis of the latent space cell representations (Figure 1C).

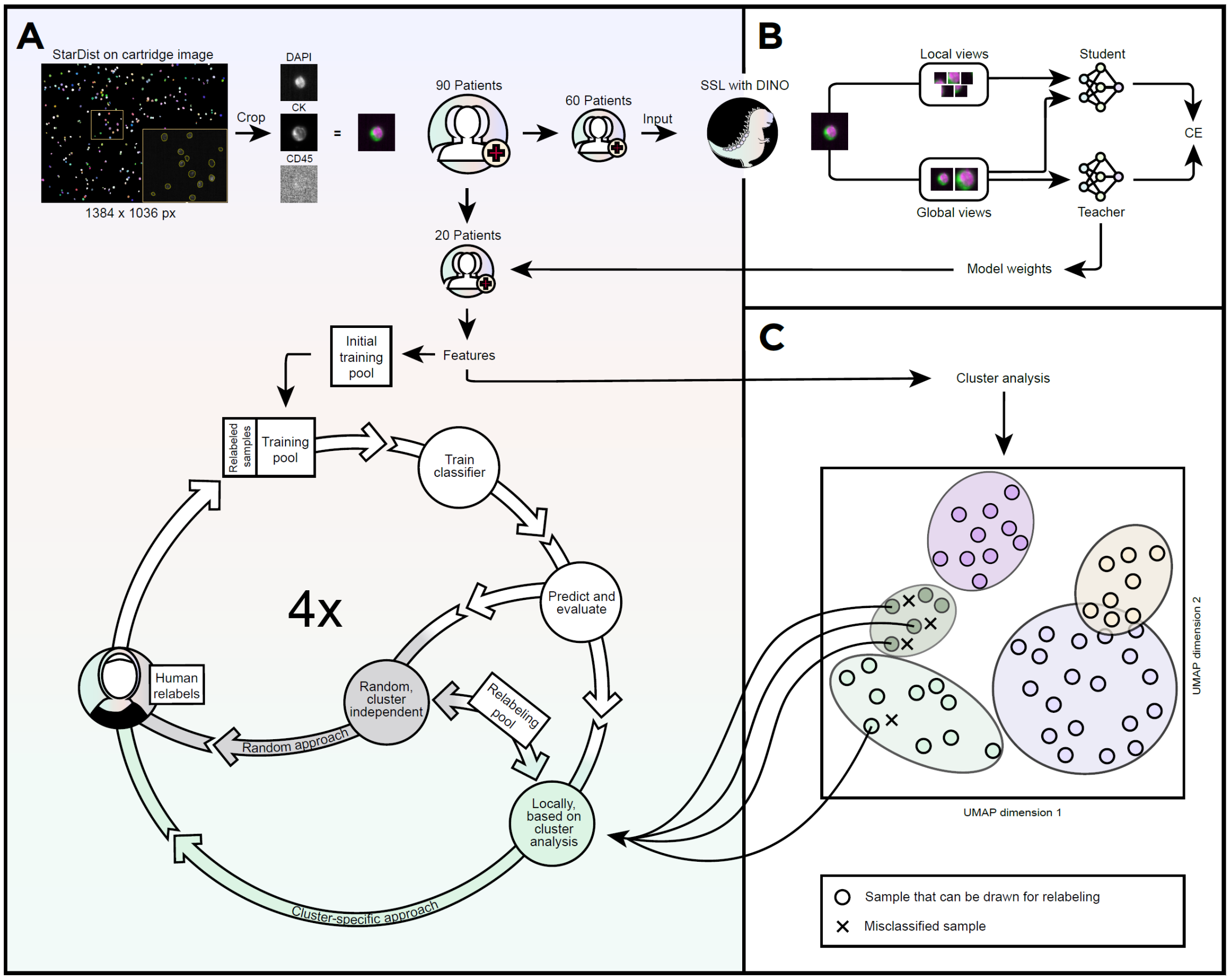

**Figure 1: Framework overview and the HiL principle.** (A) The flowchart begins with a cartridge image from the CS system, where single cells are segmented and cropped using StarDist, and the available single-channel images are merged into a three-channel image. (B) The DINO network is trained with data from 60 patients, while 20 undergo classification using a conventional machine learning classifier (support vector machine, SVM) within the HiL framework (A). After training and evaluation, additional images were sampled from a relabeling pool and labeled by a human expert to boost classification performance. This process involves the proposed cluster-based approach and random resampling as a naive baseline approach. (C) The proposed cluster-based approach uses information from a cluster analysis based on labeled data to identify the clusters in the latent space with low F1 scores. Relabeled images are then included in the training pool. The HiL loop was applied four times. The remaining 10 patients (out of the 90) are not shown here and will be used later for the final evaluation of CTC detection performance of the proposed pipeline. Abbreviations and explanations: DAPI: Nuclei stain; CK: Tumor marker; CD45: Leukocyte marker; SSL: Self-supervised learning; CE: Cross entropy; CS: CellSearch; DINO: Self-distillation with no labels; UMAP: Uniform manifold approximation and projection.

The extraction of the single cells starts with applying the StarDist algorithm[23] to segment the cells in the CK channel of the cartridge images acquired by the CS system. The segmented cells are then cropped and organized in the order of DAPI, CK, and CD45 channels to create three-channel images that define the input of subsequent deep learning systems. In the present study, this step is applied to all the image cartridges of all 90 breast cancer patients.

The self-supervised learning (SSL)-based image encoder training followed the DINO principle. The SSL part was performed using data from 60 out of 90 patients.

The learned representations of the DINO teacher backbone were then used for the addressed downstream task, that is, the classification of the single cell images (CTC vs. non-CTC). Based on the data of the 20 patients not used for the SSL part, an initial training pool as well as a test set were defined before initiating the HiL approach (see Methods part for details). The extracted features of the training set were used to fit an SVM classifier. Additionally, a designated relabeling pool was defined, consisting of samples from previously non-annotated data.

The HiL strategy for drawing samples from the relabeling pool is depicted in Figure 1C. The central idea was to target latent space clusters that showed a low classification performance, in this study the lowest F1 score, i.e., the harmonic mean of precision and recall. We contrasted this with a baseline approach where additional samples were randomly selected from the relabeling pool, independent of a cluster association. A human operator carried out relabeling of the new samples, and the labeled samples were added to the training pool to adapt the SVM decision boundary and re-evaluate the classification performance. For the performed experiments, the HiL loop depicted in Figure 1A (lower part) was applied four times.

## Cluster identification and characterization

The cluster analysis aimed at the automatic identification of areas in the latent space with a low F1 score compared to other areas, i.e., areas with relatively many misclassifications.

The results of the cluster analysis are summarized in Figure 2B. A total of five clusters with varying sizes and shapes were identified. Data points not assigned to any of these clusters were referred to as belonging to the background cluster. The meaningfulness of each cluster was then assessed, confirming that cell images within the same cluster exhibited similar characteristics. For example, cluster 1 contained cell images with rather small and point-like signals, both in the DAPI and CK channels, while cluster 2 primarily showed many DAPI signals in the background. Furthermore, cluster 0 contained images where a shine-through effect occurs, originating from a strong fluorescence signal in the CK channel that extended into the CD45 channel[24]. Additionally, we observed images displaying artifacts such as smeared cells and noisy data spread across the clusters (Figure 2A). Upon this finding, an additional ML classifier was trained to preselect these images and to include only valid cell images in the subsequent classification task.

Regarding the classification performance on the test set, we observed that each cluster contained varying amounts of misclassifications, with cluster 2 showing the highest number of misclassified cells and the lowest F1 score (see red dashed box in

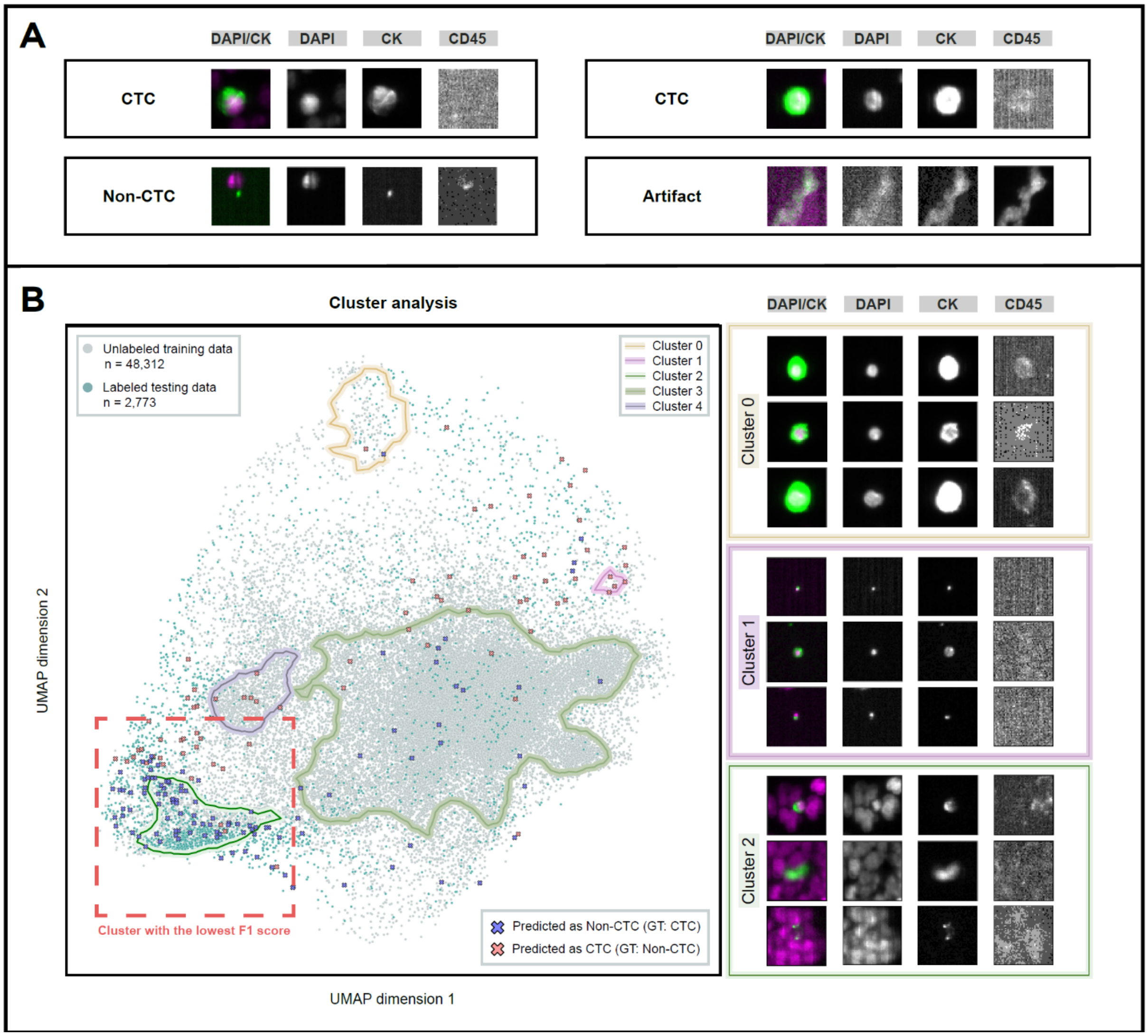

**Figure 2: Cluster identification and evaluation.** (A) Displays representative cell images with corresponding DAPI, CK, and CD45 channels, including the overlap of DAPI and CK: a CTC (positive for DAPI and CK, negative for CD45), a non-CTC, a shine-through effect from CK to CD45 channel, and an artifact example. The left part of (B) displays the latent space of trained image encoder, reduced to two dimensions by a UMAP transform, for the labeled test and unlabeled training data. Clusters identified via clustering are highlighted by closed contours. Data points not assigned to any of the identified cluster are defined as background. The cluster with the lowest F1 score, based on the labeled test data, is highlighted by a red dashed box. Misclassified cells are indicated by crosses. The right part of (B) depicts exemplary cell images from clusters 0, 1 and 2. The first two rows of the cluster examples contain cell images from the labeled test data, and the third row from the unlabeled training data. Abbreviations and explanations: DAPI: Nuclei stain; CK: Tumor marker; CD45: Leukocyte marker; GT: Ground truth.

Figure 2B). A closer inspection of cluster 2 revealed that the predominant misclassifications within and in the vicinity of cluster 2 were non-CTC predictions of cells that are determined as CTCs by human experts.

## Impact of sampling on classification performance

The central hypothesis of the proposed HiL approach was that targeted sampling and labeling of additional cell images from automatically determined latent space areas, based on the F1 scores of the local areas, results in improved local and overall classification performance compared to fully random sampling.

To investigate the hypothesis, three HiL experiments were performed: two simulated experiments in a controlled, idealized environment and one real-world experiment with a human expert who assigns labels to unseen data. All three experiments were based on the identified clusters shown in Figure 2B, left. The performance of the cluster-based strategy to complement the classifier training data set was compared against random sampling approaches. Each HiL experiment was repeated five times. The results are summarized in Figure 3A and Table 1.

### Simulated HiL scenario 1: limited global data

The first experiment assessed the classification performance of the cluster-based HiL strategy when starting with a very limited classifier training data set. The initial training pool for this experiment consisted of a subset of only 100 labeled samples randomly selected from the labeled training set. During each simulated HiL loop, 100 additional labeled cell images were sampled and added to the classifier training pool. For the baseline approach, these images were randomly sampled. For the cluster-based strategy, new samples were drawn from the clusters with a frequency inversely proportional to the cluster-specific F1 scores before the respective HiL loop.

Figure 3A shows that the cluster with the lowest initial F1 score was cluster 2 (average F1 score for five repetitions of the experiment: 0.107; evaluation based on the test data set); the other clusters start with higher F1 scores. After four HiL loops, a noticeable classification improvement for cluster 2 is depicted using the cluster-specific approach, achieving an F1 score of 0.635. In contrast, the random sampling approach to enrich the SVM training data set reached only 0.260 after four iterations. The quantitative evaluation is supported by the qualitative impression of the latent space snapshots of cluster 2 and its proximate area (Figure 3A, lower panel): After initialization, many erroneous non-CTC predictions occurred, especially in the northern region of cluster 2 while more erroneous CTC predictions appeared in the southern region. After HiL loop 4, the number of false predictions reduced, and the reduction of false non-CTC predictions was more apparent for the cluster-specific than for the random approach.

Further, in both approaches, the F1 score increased for the majority of clusters after four HiL loops (see Figure 3A), albeit with less pronounced improvements than for cluster 2. For the cluster-based sampling approach, this can be explained by the proposed sampling strategy: With an increasing F1 score for cluster 2, the probability of of drawing and labeling new samples from the other clusters increases for the available training for the individual. On average, the F1 score increased from initially 0.849 to 0.911 for the cluster-based and 0.896 for the random approach after four HiL loops.

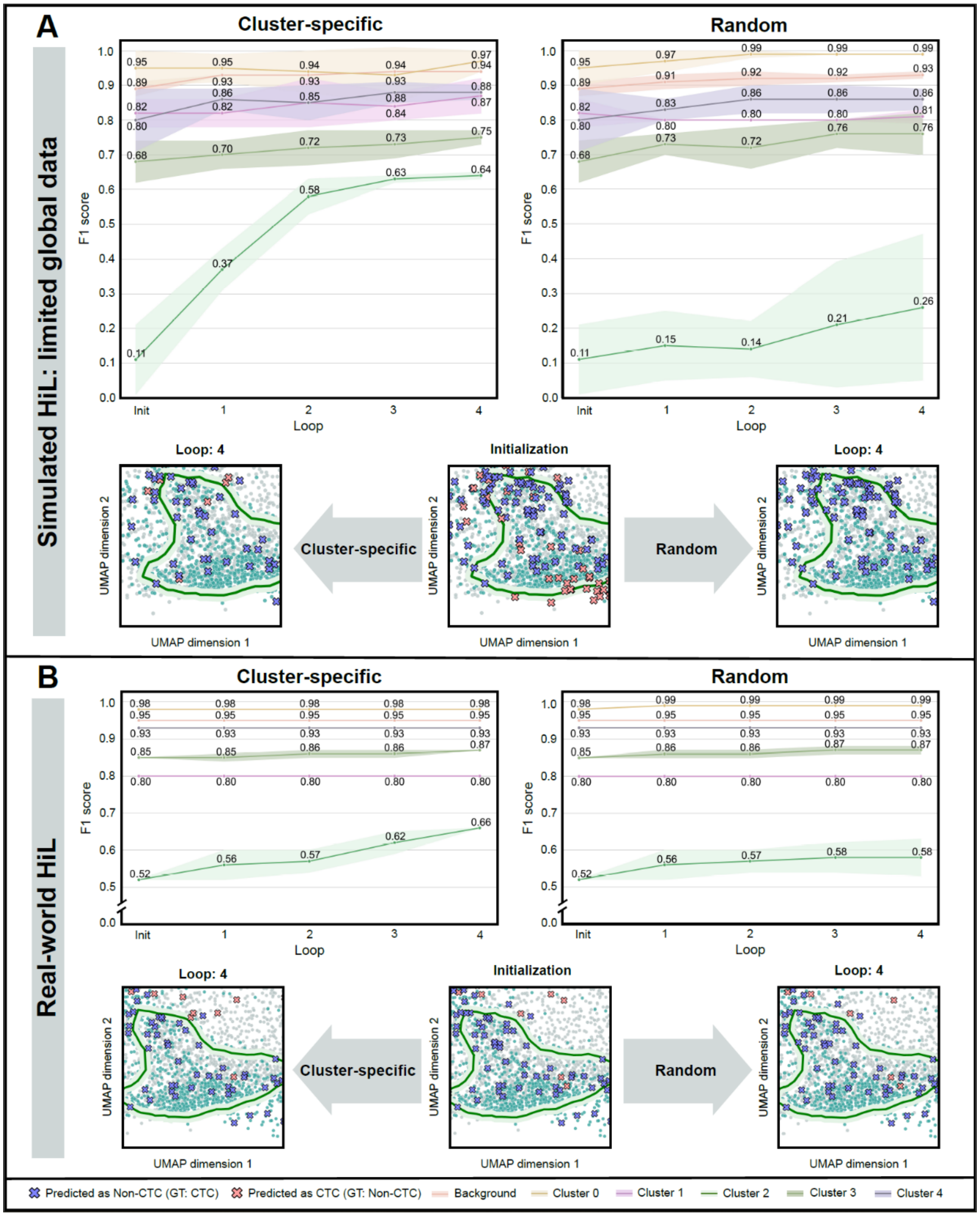

**Figure 3: Impact of HiL sampling strategy on classification performance.** Two experiment settings are depicted: Simulated sampling and relabeling (simulated HiL: limited global data) in (A) and relabeling by a human expert (real-world HiL) in (B). Line plots display mean F1 scores and standard deviations of the respective loops across the five HiL runs for each cluster, including background. Snapshots depict the latent space after initialization and final loop 4, focusing on cluster 2, i.e., the cluster with the most misclassifications, highlighting differences in prediction accuracy. Abbreviations and explanations: HiL: Human-in-the-loop; Init: Initialization; Loop 1-4: Sampling and relabeling loops.

### Simulated HiL scenario 2: limited local data

While the first experiment focused on challenges due to limited labeled training data in general, the second experiment addressed the challenge of limited training data for a specific local latent space area, i.e., a specific cluster. As the latent space captures the learned representation of the input images, this scenario corresponds to the situation that the images and cell representations of a new patient do not match the characteristics of the majority of the images and patients used for classifier training.

To mimic this scenario, the number of training samples was cut to 20% of the original training data set size for one cluster (subsequently called the main cluster) and to 80% for the other clusters. Within each loop of the simulated HiL scenario, an additional 20% of the main cluster training samples were added to the SVM training data set. As a comparison benchmark, the additional training samples were randomly drawn from the left-out samples of the other clusters and 20% of the left-out samples of the main cluster.

The experiments focused on cluster 2 (cluster with the most misclassification) and cluster 3 (largest cluster) as main clusters. The results are summarized in Table 1.

**Table 1:** Classification performance: cluster-specific vs. random sampling approach for the simulated HiL experiment 2: limited local data. Data reported are mean and standard deviation across clusters, including background. Total refers to the F1 score evaluated for the complete test set. Abbreviations: Init: Initialization; HiL: Human-in-the-loop.

| | Simulated HiL: limited local data | | | |
|---|---|---|---|---|
| | Main cluster: cluster 2 | | Main cluster: cluster 3 | |
| F1 score | Cluster-specific | Random | Cluster-specific | Random |
| Total – Init | 0.919 ± 0.003 | 0.919 ± 0.003 | 0.909 ± 0.003 | 0.909 ± 0.003 |
| Total – loop 4 | 0.921 ± 0.002 | 0.919 ± 0.003 | 0.921 ± 0.003 | 0.916 ± 0.003 |
| Background – Init | 0.945 ± 0.001 | 0.945 ± 0.001 | 0.942 ± 0.002 | 0.942 ± 0.002 |
| Background – loop 4 | 0.946 ± 0.001 | 0.945 ± 0.002 | 0.946 ± 0.002 | 0.946 ± 0.001 |
| Cluster 0 – Init | 0.982 ± 0.006 | 0.982 ± 0.006 | 0.985 ± 0.006 | 0.985 ± 0.006 |
| Cluster 0 – loop 4 | 0.982 ± 0.006 | 0.985 ± 0.006 | 0.980 ± 0.005 | 0.980 ± 0.005 |
| Cluster 1 – Init | 0.800 ± 0.000 | 0.800 ± 0.000 | 0.818 ± 0.040 | 0.818 ± 0.040 |
| Cluster 1 – loop 4 | 0.800 ± 0.000 | 0.800 ± 0.000 | 0.808 ± 0.019 | 0.800 ± 0.000 |
| Cluster 2 – Init | 0.432 ± 0.087 | 0.432 ± 0.087 | 0.315 ± 0.082 | 0.315 ± 0.082 |
| Cluster 2 – loop 4 | 0.492 ± 0.056 | 0.456 ± 0.109 | 0.485 ± 0.079 | 0.423 ± 0.029 |
| Cluster 3 – Init | 0.854 ± 0.012 | 0.854 ± 0.012 | 0.799 ± 0.016 | 0.799 ± 0.016 |
| Cluster 3 – loop 4 | 0.847 ± 0.018 | 0.847 ± 0.018 | 0.852 ± 0.017 | 0.809 ± 0.027 |
| Cluster 4 – Init | 0.922 ± 0.015 | 0.922 ± 0.015 | 0.895 ± 0.024 | 0.895 ± 0.024 |
| Cluster 4 – loop 4 | 0.919 ± 0.009 | 0.919 ± 0.009 | 0.919 ± 0.009 | 0.919 ± 0.009 |

With cluster 2 as the main cluster, the cluster-specific HiL approach yielded a higher F1 score (0.492 for cluster 2 after four HiL loops) than the random sampling strategy (F1: 0.456). This trend was consistent for the entire testing set evaluation (cluster-specific HiL: F1 score of 0.921; random sampling: 0.919).

With cluster 3 as the main cluster, the cluster-specific HiL strategy surpassed random sampling for both the main cluster (cluster-specific HiL: F1 score of 0.852, random sampling: 0.809) and the neighboring cluster 2 (cluster-specific HiL: 0.485, random HiL: 0.423).

### Real-world HiL experiment

For the real-world experiment, the initial classifier training pool was the entire labeled training set. Additional samples for classifier refinement were drawn from unseen and unlabeled data, and the new samples were labeled by a human expert. The experiment focused again on cluster 2. Since most of the initial misclassifications in cluster 2 were erroneous non-CTC predictions, only new samples from cluster 2 that were classified as non-CTCs were considered for expert labeling. To enrich the SVM training dataset, only the subset of these samples categorized as CTCs by the human expert was used. The labeling time was limited to 5 minutes per loop (see Methods for further details).

The results are summarized in Figure 3B. During the labeling periods, the expert identified in total 32 CTCs in the initially unlabeled samples of cluster 2 that were erroneously classified as non-CTCs. Albeit the small number of additional new SVM training samples, the proposed HiL strategy resulted in an increase in the F1 score for cluster 2 from initially 0.524 to 0.661 after four HiL loops, illustrating the efficacy of the proposed targeted sampling strategy. A similar random sampling strategy led to an F1 score of 0.578 after four loops. This trend is further evident in the latent space. Snapshots of cluster 2 and its surroundings reveal relatively fewer misclassifications after the last loop for the cluster-specific strategy than for the random one (see Figure 3B, bottom). Small improvements in the F1 score were also noted in the neighboring clusters, such as cluster 3. Further, starting from an overall F1 score of 0.923 for the complete test set, the cluster-specific approach achieved a higher F1 score (0.930) than the random one (0.926).

## Application of final model

To evaluate the CTC detection performance of the proposed training strategy, the final model of the cluster-specific real-world HiL experiment was applied to 10 additional patients. The performance of the proposed pipeline was compared to the CS system in terms of the number of identified CTCs and the positive predictive value. The latter is defined by how many of the cells and events, respectively, shown to the human observer are actual CTCs. The results are summarized in Table 2.

**Table 2:** Summary of the CTC detection results of the proposed HiL system and the CS system. The positive predictive value indicates the fraction of (proposed) cells that are actual CTCs. Events refer to the images presented in the CS gallery.

| | Final HiL model | | | CS | | |
|---|---|---|---|---|---|---|
| Patient | Suggested CTC candidates | Actual CTCs | Positive predictive value | Events | Actual CTCs | Positive predictive value |
| 1 | 91 | 77 | 0.846 | 239 | 74 | 0.310 |
| 2 | 135 | 104 | 0.770 | 531 | 107 | 0.202 |
| 3 | 91 | 76 | 0.835 | 219 | 80 | 0.365 |
| 4 | 38 | 24 | 0.631 | 101 | 23 | 0.228 |
| 5 | 225 | 124 | 0.551 | 1197 | 113 | 0.094 |
| 6 | 71 | 34 | 0.479 | 168 | 38 | 0.226 |
| 7 | 50 | 31 | 0.620 | 133 | 34 | 0.256 |
| 8 | 23 | 20 | 0.870 | 167 | 20 | 0.120 |
| 9 | 23 | 14 | 0.609 | 93 | 16 | 0.172 |
| 10 | 148 | 130 | 0.878 | 790 | 143 | 0.181 |

The overall number of actual CTCs identified across the 10 patients was comparable for both systems, whereas the positive predictive value of the proposed pipeline was noticeably higher for all patients, resulting in a lower number of false positive images that need to be analyzed. This, in turn, is less time-consuming for the human expert.

In a subsequent analysis, the actual CTCs identified by both systems were examined to assess the overlap in CTC detection between the two systems and to identify any CTCs detected by only one system. For many patients, some CTCs suggested by the CS system were not detected by the proposed system (see Figure 4B) (CTCs found only by CS across patients: range: 2-18; average: 7), and vice versa (CTCs found only by proposed system: range: 0-26; average: 6). Exemplary CTCs found by the model but not by the CS system, are depicted in Figure 4A. Among these, there are CTCs with a relatively lower DAPI signal intensity (see the third CTC; A) or lower CK signal intensity (see the first and fourth CTC; A). Further, the CTC detected in the second row (A) exhibits a small DAPI signal but overlaps with the CK signal.

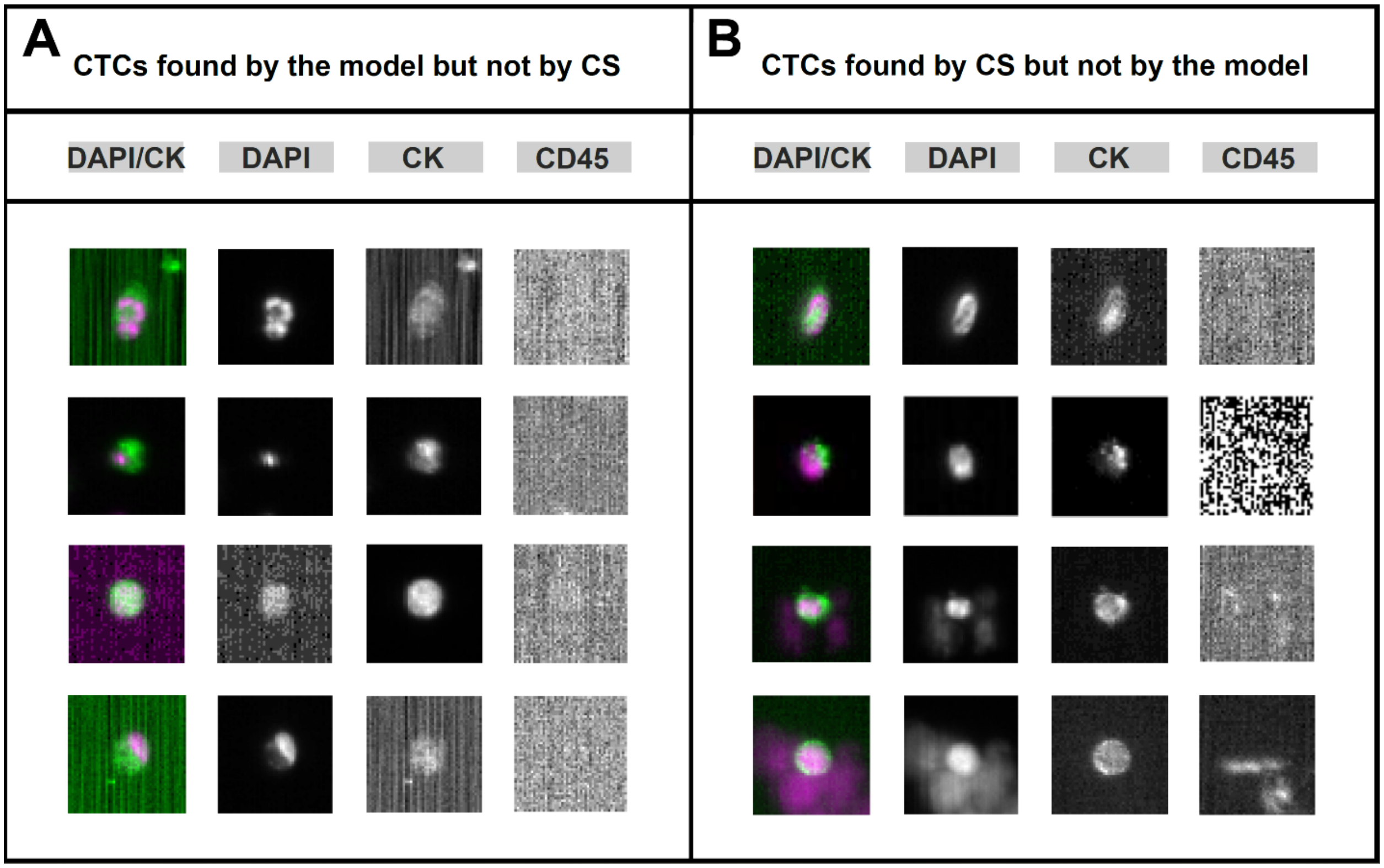

**Figure 4: Comparison of CTCs detected using the proposed model and the CS system.** (A) CTCs found by the model, but not by CS; (B) vice versa. Abbreviations and explanations: DAPI: Nuclei stain; CK: Tumor marker; CD45: Leukocyte marker; CS: CellSearch.

# Discussion

In recent years, multiple efforts have been made to automate the detection of CTCs. These include the application of machine learning techniques, encompassing supervised learning approaches[17], semi-supervised methods[19], and, taking up current trends in ML, for instance first self-supervised techniques[21]. Many of these developments address existing limitations posed by the FDA-cleared CS system, which is regarded as the reference and gold standard in this field: Due to the semi-automated nature of the CS system, it requires the intervention of a skilled human operator to select CTCs from sometimes a large number of images, introducing a time-intensive aspect to the process.

While we acknowledge the efforts that aim at complete automation of CTC detection, we argue that leveraging human expertise will remain crucial, especially for the clinical application of such systems. We think that human input will remain essential especially in cases where ML system uncertainties arise, for instance, due to a mismatch of the training population and the specific patient and blood samples to be analyzed. However, in such cases, human expert input is also valuable to further optimize the ML system predictions, resulting in a human-in-the-loop (HiL) scenario.

In this study, we introduced a novel HiL strategy that bridges the gap between self-supervised and semi-supervised methodologies. We combined a self-supervised DL feature extraction with a conventional ML classifier to perform a binary classification of CTCs and non-CTCs. Leveraging self-supervised feature extraction enabled us to learn comprehensive cell representations using a large amount of unlabeled data that is always available in related clinical settings. Using only a limited amount of labeled data enabled the identification of clusters in the latent space with low classifier performance. The latent space analysis then allowed us to generate (pseudo) labels for uncertain regions; to focus human efforts on labeling a limited set of new samples for classifier improvement only where they are most needed; and, thereby, to improve the classifier predictions and increase the ML system certainty.

The feasibility and the advantages of the proposed strategy were demonstrated for liquid biopsy data of metastatic breast cancer patients. We showed the existence of distinct latent space clusters of cells of similar characteristics (shape, size, feature expressions) and observed differing system classification performances for the different clusters. The proposed iterative cluster-specific HiL strategy was compared to a random sampling approach that was independent of cluster associations of newly sampled data points. Across all our experiments, encompassing real-world HiL and simulated HiL experiments, we consistently observed a faster increase in classification accuracy in overall performance as well as in local performance, particularly for the cluster initially displaying the lowest F1 score, when employing the cluster-specific approach. Furthermore, the observed total classification accuracy of our framework was higher than the corresponding numbers reported by Nanou et al.[19] for their testing set from metastatic breast cancer (precision on CTC: 0.938 vs. 0.814, recall on CTC: 0.923 vs. 0.793). Compared to CS, we showed that the proposed pipeline enables identification of a similar number of CTCs for patients that were not part of the training and optimization of the system but with a much smaller image gallery presented to the observer, i.e., a smaller number of proposed cell images and events that needs to be evaluated by the human expert. This, in turn, saves the time of the expert.

From a clinical application perspective, the proposed HiL strategy in combination with latent space analysis enables the identification of latent space areas with higher classification uncertainty and proposes a targeted strategy to reduce uncertainty. This means that for each blood draw, the user can see whether the system accuracy can be expected to be high for the respective latent space cluster. If this is not the case, the user can ask the system to suggest potential CTC candidates for visual inspection that, based on the classifier probability threshold, are considered less certain CTC candidates. In the next step, the results of the expert evaluation can be fed back into the system to improve the classification performance for the cluster - and thus save time in the long term when again blood samples with similar characteristics are to be analyzed.

## Limitations of the study

The current study focused on the feasibility and proof-of-concept of the proposed HiL strategy. The feasibility was demonstrated using liquid biopsy data from metastatic breast cancer patients. We further showed that the CTC detection performance of the proposed pipeline is comparable to that of the CS while presenting less images to the human operator. CTCs detected by our approach but not by the CS system and vice versa may be attributed to the different thresholds applied by the two systems to the cartridge images. However, the exact reasons why one system detects a CTC when the other does not remain to be investigated.

We propose expanding the scope of our present study to include other tumor entities, especially metastatic prostate or colon cancer, for which the CS system has also been approved. Further, so far, we used data with high CTC counts, which emphasizes the need to extend and evaluate our approach to patient data with lower reported CTC numbers and data of healthy blood donors. This is particularly important considering the predictive significance attributed to the current consensus threshold of 5 CTCs per 7.5 ml blood draw for both progression-free survival and overall survival[13]. The potential of the proposed framework in a comprehensive clinical setting has, therefore, to be investigated as future work.

# Experimental procedures

## Resource availability

### Lead contact

Hümeyra Husseini-Wüsthoff is the lead contact of this study and can be reached via e-mail (h.husseini-wuesthoff@uke.de).

### Materials availability

This study did not generate new unique reagents.

### Data and code availability

The processed data supporting the cluster analysis and the simulation findings, and the model weights for the DINO image encoder are available at Zenodo https://zenodo.org/records/14033379. The source code is accessible on GitHub https://github.com/IPMI-ICNS-UKE/CTC-HiL. Image data can be shared upon request by contacting the lead contact.

## Materials: data description

### Liquid biopsy data preparation

This study is based on liquid biopsy data of 90 metastatic breast cancer patients, i.e., cartridge images obtained through the CS system. Cartridge images were obtained by transferring enriched and stained cells from a 7.5 ml whole blood sample into a cartridge, which is then subjected to a magnetic field to pull the cells in a single focal plane against a glass surface[16]. A fluorescence microscope then scans the cartridge and creates 175 digitized cartridge images with a size of 1384 x 1036 px. Each cartridge image consists of three channels, representing the three applied staining agents: DAPI, CK and CD45.

CTCs were detected in blood samples from patients with metastatic breast cancer (mBC) treated at the University Medical Center Heidelberg, Germany. CTC counts were determined by the CellSearch® approach in the Institute of Tumor Biology, University Medical Center Hamburg-Eppendorf, Germany. CTC-analyses were approved by the ethical committee of the University of Heidelberg (case numbers S295/2009 and S-164/2017; NCT05652569), and University of Mannheim (2010-024238-46) and by the ethic committee of the chamber of physicians of Hamburg (5392-3704-BO) and all patients provided their written consent.

For detection and segmentation of single cells in the CS cartridge images, StarDist[23] was applied, which has already been shown to be well suited for single cell segmentation in similar context[25]. In this study, the cells were detected and segmented based on the CK channel[21]. Based on the segmentation information, cropped three-channel single cell images of size 48 x 48 px were generated. Further preprocessing (min-max intensity normalization to a channel intensity range between 0 and 1) followed the protocol described by Husseini et al.[21].

### Data split

The processing and cell segmentation of the cartridge images of the 90 patients led to a total of 1.321.951 (three-channel) single cell images. 60 patients with 999,285 cell images were used for self-supervised training of the image encoder. 20 patients (274,542 cells) were used for solving the downstream task of binary classification of cells into CTCs and non-CTCs and respective experiments. 5,411 cells were labeled by two domain experts and evaluated by consensus, resulting in 2,773 CTCs and 2,688 non-CTCs. The 20 patients were randomly split into 10 training (GT: 1,509 CTCs, 1,129 non-CTCs) and 10 test patients (GT: 1,214 CTCs, 1,559 non-CTCs). The unlabeled cell images of the training patients were used as an unlabeled cell image pool for sampling and labeling additional cell images during the HiL experiments.

Due to the presence of noisy images within the unlabeled training data set, i.e., images with channel signals that are hardly interpretable for the human observer, an SVM was trained on a small subset of the unlabeled images (400 images that were considered noisy by the human experts; 400 that were not noisy) to identify such samples. The trained

SVM was applied to the entire unlabeled training data set and noisy images were excluded. The refined and final unlabeled training data set comprised 48,312 samples.

The remaining 10 patients with 48.124 cell images were set aside for the final evaluation of CTC detection performance of the proposed pipeline, compared to the CS system.

## Methods

### Self-supervised image encoder training

Self-supervised DL image encoder training was based on the unlabeled single cell three-channel images of 60 patients (999,285 images) using the public *sparsam* implementation[20] of the DINO framework by Caron et al.[22].

Based on the concept of knowledge distillation between two DL models[26] and building up on contrastive learning frameworks and momentum encoders[27], the DINO framework consists of a teacher and a student DL model, both sharing the same architecture while being trained on different patch views of the same input image. As shown in Figure 1B, an input image is randomly cropped into two global and five local crops. The teacher receives only the global crops while the student obtains all crops. The objective of the training is to minimize a temperature-weighted categorical cross entropy between student and teacher model outputs, thereby learning a consistent representation of the different patches of the same input image. The student model parameters are optimized by stochastic gradient descent and the teacher model parameters are computed as running exponential mean of the student parameters[22].

Each model consists of a backbone network and a projection head that is only used during SSL training. The setup in this study follows the one of Nielsen et al.[20] and Husseini et al.[21] and deploys a cross-variance vision transformer (XCiT)[28] as the backbone. Training was carried out for 30,000 iterations. After training, the teacher model backbone was applied to infer the image representations used for latent space analysis and image classification.

### SSL feature-based image classification

Following the recent success of combining SSL-trained DL models and standard ML classifiers for image analysis for scenarios with only a few annotated data[20, 21], an SVM with default scikit-learn parameters (except for: class weight=binary, cache size=10,000, probability=True, and breakties=True) was applied for the sought binary image classification (CTC vs. non-CTC) task. The input to the SVM was the representation of the single cell image as extracted by the trained teacher backbone model, which was further reduced from 128 to 32 dimensions by principle component analysis (PCA). The SVM was fitted to the labeled training data set.

## Latent space cluster analysis

The latent space cluster analysis was performed based on the PCA-reduced SSL image representations that were also used as inputs of the SVM. Following common practice[29, 30] a Uniform Manifold Approximation and Projection (UMAP) was further applied to obtain a two-dimensional representation of the latent space. Clustering in this two-dimensional space was carried out using Hierarchical Density-Based Spatial Clustering of Applications with Noise (HDBSCAN). HDBSCAN can automatically determine a suitable number of clusters and identify clusters with varying densities and shapes (e.g., compared to Density Based Spatial Clustering of Applications with Noise (DBSCAN)[31]). Both properties were desirable in the present study, as little was known about the true data distribution, such as the true number of clusters and their characteristics.

The cluster analysis was performed on the UMAP features of the joint unlabeled training and the test data set. While this might seem unintuitive at first glance, it serves an important purpose: to study local cluster effects in a real-world scenario (e.g., the real-world HiL experiment), non-training data must be assigned to a cluster; therefore, test data must be included in the clustering process. Furthermore, to compensate for the limited size and potential biases when using only the labeled part of the test data set, we enriched the test set with the much larger and potentially more diverse unlabeled training set. During evaluation, data samples not assigned to any cluster by HDBSCAN were referred to as the background data.

## Proposed HiL strategy

The underlying idea of the proposed HiL strategy was to improve labeling efficiency and classification performance by targeted automatic sampling and labeling new data points from latent space clusters with low(er) classification performance. The iterative process consists of the following steps:

1. **Initialization:** A pool of unlabeled samples is defined, called relabeling pool, along with an initial training pool of labeled samples. Furthermore, a test set is prepared for evaluation purposes.

2. **Relabeling loop:** The following steps constitute the loop and are repeated until the classification performance is satisfying or no more data are available in the relabeling pool:

   i. The classifier (SVM) is fitted to the labeled training samples.
   ii. The fitted classifier is applied to classify the labeled cell images of the test set.

iii. The local performance of the classification is evaluated for the latent space clusters and the labeled test set. In this study, the F1 score was used and computed for each cluster and the entire test set.

iv. Cluster-specific sampling of new data points: Based on the classification performance for the different clusters, unlabeled samples are drawn from the relabeling pool.

v. The drawn samples are presented to a human expert who assigns class labels. The labeled images are added as new data points to the training pool and removed from the relabeling pool.

## Experiments

The working hypothesis of the proposed HiL strategy was that it allows improving classification performance with fewer new labeled samples compared to a naive, fully random sampling approach. The hypothesis was tested in two simulation experiments (simulated HiL scenarios 1 and 2, performed in an idealized environment) and a real-world implementation of the strategy, involving a human expert and unseen new data (real-world HiL experiment).

Each experiment comprised four relabeling loops and was repeated five times with different random seeds to study the robustness of the results. The classification performance was assessed by the F1 score for the different clusters and the overall F1 score.

### Simulated HiL scenario 1: limited global data

In the first scenario, classification improvement through local training data set adaptation was assessed when starting with very limited initial training data. For this simulation experiment, only 100 labeled samples were selected randomly from the labeled training pool to form the initial SVM training data set. The remaining labeled cells of the original training data set defined the relabeling pool for this experiment. During each relabeling loop, 100 additional samples were drawn from this pool, resulting in 500 samples after one completed HiL experiment (i.e., after four loops). During each loop, a 100-fold Monte Carlo (MC) cross-validation of the limited labeled training pool of this experiment was performed by splitting the data into MC-training (90%) and MC-validation (10%) for each step. All MC-validation results were then combined, and the classification performance of each cluster was evaluated in terms of the F1 score. For cluster-specific sampling, new samples were drawn from the relabeling pool and the clusters with a relative frequency $c_i$ inversely proportional to the associated MC-validation F1 score $s_i$ of cluster $i$,

$$c_i = \frac{1 - s_i}{\sum_j 1 - s_j}$$

As a baseline comparison, a random sampling approach was simulated, where in each loop 100 new samples were randomly drawn from the entire relabeling pool of this experiment, agnostic of cluster performance.

### Simulated HiL scenario 2: limited local data

The second experiment emphasized the scenario of limited labeled training data for a specific local area in the latent space in the initialization phase. For this scenario, the training data set of only a single cluster (referred to as the main cluster) was pruned to 20% of its original size, while all others were limited to 80% of their original size. Due to the limited number of training samples for SVM fitting, the classification performance for the main cluster was hampered compared to the other clusters. During each relabeling loop, the next 20% of the labeled original training data set of the main cluster were randomly drawn (calculated based on 100% of the total available samples from the main cluster) until 100% (i.e., all labeled samples) of the main cluster were used for classifier fitting.

For comparison purposes, the same initial labeled training set was defined, but we used 20% of the left-out labeled data for each cluster, including the main cluster, to form the relabeling pool. Sampling from the relabeling pool was then carried out randomly, with the number of drawn samples per loop given by the corresponding number for the cluster-specific sampling strategy.

This experiment was only conducted for the two most interesting clusters: cluster 2, which represented the cluster with the lowest F1 score at initialization, and cluster 3, which represented the largest cluster.

### Real-world HiL experiment

The real-world HiL experiment followed the same scheme as the simulated HiL but did not rely on artificially limited labeled classifier training data. Instead, new training data were sampled from initially unlabeled samples. That is, the initial classifier was trained on the entire labeled training data set, and the test set was used to evaluate the cluster-specific classification performance.

As for the relabeling pool, 1000 new samples were drawn from the unlabeled training set. We focused on the cluster with the initially lowest F1 score (cluster 2) and constrained the relabeling pool by sampling only from this cluster. For cluster 2, the low F1 score was mainly due to erroneous non-CTC prediction. We therefore limited the relabeling pool to samples for which the classifier predicted the cell to be non-CTC. Interested in mainly reducing the number of erroneous non-CTC predictions, the human expert was given 5 minutes to identify as many of these false predictions as possible, taking observer variability into account[32]. These were then, as new CTC examples, added to the classifier training pool.

In the first HiL run, 32 new samples were labeled as CTC, with 11 samples in loop 1, 10 samples in loop 2, 7 samples in loop 3, and 4 samples in loop 4. Since no further

CTCs were found, the new labeled samples were shuffled in the remaining repeated HiL runs, while maintaining the same number of new samples per loop.

For the random sampling approach, also only non-CTC predictions from the unlabeled training set were sampled, but without considering their cluster association. Furthermore, only images that were initially predicted to be non-CTCs but were identified as CTCs by the expert were added to the training pools to match the cluster-specific experiment design. Furthermore, the number of added samples per loop was the same as for the cluster-specific experiment.

### Application of final model

To assess the performance of the proposed HiL strategy for CTC detection, the final SVM model of the real-world cluster-specific HiL experiment (after four HiL loops) was applied to the remaining 10 patients who had not been used in the aforementioned experiments. Similar to the other experiments, the input for the SVM consisted of the PCA-reduced representations of single-cell images extracted by the trained teacher backbone model. CTC candidates were determined utilizing the SVM decision function with a confidence threshold of 0.5 (minor adjustments to 0.4 and 0.6 showed similar positive predictive values), and duplicate cells were automatically removed.

Similar to the image gallery of the CS system, the suggested CTC candidates were presented to an expert, and CTCs were identified. For all patients, the same expert analyzed the CS image gallery and identified the CTCs therein. The number of CTCs was counted for both systems. For the identified CTCs, their coordinates in the CS cartridge image were extracted from the Extensible Markup Language (.xml) file generated by CS, and the fraction of CTCs that were identified in the image galleries of both systems was analyzed. Cells that were labeled as CTCs for one system and as non-CTCs for the other were taken into account during evaluation.

## Acknowledgments

This work is funded by the Erich und Gertrud Roggenbuck-Stiftung. SR and KP received funding from the Deutsche Krebshilfe (Förderschwerpunktprogramm ‘Translationale Onkologie’; Grant 70114705). The authors thank Sara Tiedemann for assistance with writing the manuscript.

## Author contributions

Conceptualization, H.H-W., M.N., and R.W.; Validation, H.H-W. and M.N.; Resources, S.R., A.S., A.T., K.P., H.W., and R.W; Writing – Original Draft, H.H-W.; Writing – Review & Editing, H.H-W., S.R., A.S., A.T., K.P., H.W., M.N., and R.W.; Visualization, H.H-W.;

Supervision, S.R., M.N., and R.W.; Project administration, R.W.; Funding Acquisition, S.R., K.P., H.W., and R.W.

## Declaration of interests

The authors declare no competing interests.